# MetaOpenFOAM: an LLM-based multi-agent framework for CFD

Yuxuan Chen[a], Xu Zhu[a], Hua Zhou[a], Zhuyin Ren[a*]

[a] *Institute for Aero Engine, Tsinghua University, Beijing 100084, China*

\* *Corresponding author:* zhuyinren@tsinghua.edu.cn

**Abstract**

Remarkable progress has been made in automated problem solving through societies of agents based on large language models (LLMs). Computational fluid dynamics (CFD), as a complex problem, presents unique challenges in automated simulations that require sophisticated solutions. MetaOpenFOAM, as a novel multi-agent collaborations framework, aims to complete CFD simulation tasks with only natural language as input. These simulation tasks include mesh pre-processing, simulation and post-processing, etc. MetaOpenFOAM harnesses the power of MetaGPT's assembly line paradigm, which assigns diverse roles to various agents, efficiently breaking down complex CFD tasks into manageable subtasks. Langchain further complements MetaOpenFOAM by integrating Retrieval-Augmented Generation (RAG) technology, which enhances the framework's ability by integrating a searchable database of OpenFOAM tutorials for LLMs. Tests on a benchmark for natural language-based CFD solver, consisting of eight CFD simulation tasks, have shown that MetaOpenFOAM achieved a high pass rate per test (85%), with each test case costing only $0.22 on average. The eight CFD simulation tasks encompass a range of multidimensional flow problems, covering compressible and incompressible flows with different physical processes such as turbulence, heat transfer and combustion. This demonstrates the capability to automate CFD simulations using only natural language input, iteratively correcting errors to achieve the desired simulations at a low cost. An ablation study was conducted to verify the necessity of each component in the multi-agent system and the RAG technology. A sensitivity study on the randomness of LLM showed that LLM with low randomness can obtain more stable and accurate results. Additionally, MetaOpenFOAM owns the ability to identify and modify key parameters in user requirements, excels in correcting bugs when failure match occur, and enhances simulation capabilities through human participation, which demonstrates the generalization of MetaOpenFOAM.

**Keywords**: large language models, multi-agent system, computational fluid dynamics, retrieval-augmented generation, automated simulation

# 1. Introduction

Computational Fluid Dynamics (CFD) is a discipline that uses numerical methods and physical models to solve fluid mechanics problems [1]. Since the introduction of CFD, scientists and engineers have employed complex code to simulate and predict fluid behavior, with open-source software like OpenFOAM being a notable and mature example [2]. Although OpenFOAM has been successfully applied in many fields [3-6], it still requires researchers to possess high-level programming and specialized skills. As technology progresses, automated tools and user-friendly interfaces have emerged, allowing users to perform complex CFD simulations with simply clicking buttons on GUI, leading to the development of industrial software like Fluent [7] and COMSOL [8]. However, even with these improvements, conducting CFD simulations remains highly technical, requiring specialized knowledge and substantial manual operations. Recently, the rapid advancement of natural language processing (NLP) technologies, particularly the advent of large language model (LLM) [9-14], has brought new hope to CFD research, promising to revolutionize the field.

The emergence of LLM represents a significant breakthrough in natural language processing. LLM can understand and generate natural language, handling vast amounts of information and providing intelligent feedback. Recent advancements, such as GPT-4 [15], Llama 2 [16], and ChatGLM [17], have demonstrated powerful capabilities in tasks like translation, text generation, and question-answering. However, despite their impressive performance in many tasks, single LLM still face limitations in solving complex problems requiring extensive text generation [10, 13, 14, 18-20]. For example, CFD problems typically involve intricate geometric modeling, physical modeling, and numerical methods, which exceed the current capabilities of individual LLM. To harness the potential of LLM in the CFD field, new approaches are needed to enhance their ability to tackle complex problems.

To address the limitations of single LLM in solving complex problems, the development of Multi-Agent System (MAS) has emerged as a promising approach [9-14, 20]. MAS involves the collaboration of multiple intelligent agents to complete complex tasks, with each agent focusing on different sub-tasks or domains, thereby improving the overall system's efficiency and accuracy. Notably, MAS can achieve unsupervised adversarial generation by iteratively having certain agents evaluate the text generated by other agents, helping refine and enhance the precision of the generated text to better meet user requirements. In CFD simulation tasks, MAS can assign different agents to handle CFD task division, input file writing, CFD simulation, and simulation result evaluation. The evaluating agents can provide feedback to other agents, optimizing their outputs. Once implemented, this natural language-based MAS approach to CFD simulation can significantly lower the barrier to conduct CFD simulations and reduce the workload for researchers, making CFD studies more efficient and accessible.

Currently, various LLM-based tools have been developed to facilitate the collaboration and integration of multi-agent systems. MetaGPT [11] and Langchain [21] are two notable tools in this field. MetaGPT is a framework designed for multi-agent collaboration, enabling the coordination of multiple LLM agents to tackle complex CFD problems. Through MetaGPT, researchers can chain different LLM agents together, each playing a specific role, to accomplish tasks collectively. Langchain, on the other hand, is a tool for Retrieval-Augmented Generation (RAG) technology. By effectively integrating information from multiple documents, Langchain can provide more professional support for CFD research. Building on the MetaGPT and Langchain frameworks, this paper develops MetaOpenFOAM, a natural language-based CFD simulation framework. MetaOpenFOAM takes user requirements as input, generates OpenFOAM input files through LLM, and returns simulation results that meet user requirements after automatically running OpenFOAM.

The structure of this paper is as follows: first, the basic framework of MetaOpenFOAM and the implementation method of RAG are introduced. Next, a series of test datasets and evaluation methods for CFD simulations with natural language input are presented. Following this, the results of MetaOpenFOAM are quantitatively introduced, along with an ablation study and parameter sensitivity analysis. Finally, the results of MetaOpenFOAM are qualitatively analyzed.

## 2. Methodology

### 2.1 MetaOpenFOAM Framework

As shown in Figure 1, MetaOpenFOAM is architected to interpret user requirements, decompose them into manageable subtasks, and execute these subtasks through a series of specialized agents. The framework leverages the MetaGPT assembly line paradigm to assign distinct roles to each agent, ensuring efficient task execution and error handling.

The framework is divided into four primary roles, each with specific responsibilities and actions:

**Architect (Role)**: This role is responsible for the initial interpretation of the user's natural language requirements. The Architect converts user requirements into a specified format, finds similar cases from the database, creates the input architecture, and oversees the overall workflow to ensure that the user's specifications are accurately translated into actionable tasks.

**Actions of Architect**:

Find similar cases in the database of OpenFOAM cases and tutorials.

Create the input architecture according to the similar case.

**InputWriter (Role)**: The InputWriter role focuses on generating and refining the necessary input files for the CFD simulation. This involves writing initial input files, rewriting them as necessary based on feedback or errors, and ensuring that all files conform to OpenFOAM's requirements.

**Actions of InputWriter**:

Write input files for OpenFOAM based on the architecture provided by the Architect.

Rewrite input files for OpenFOAM if modifications or corrections are needed.

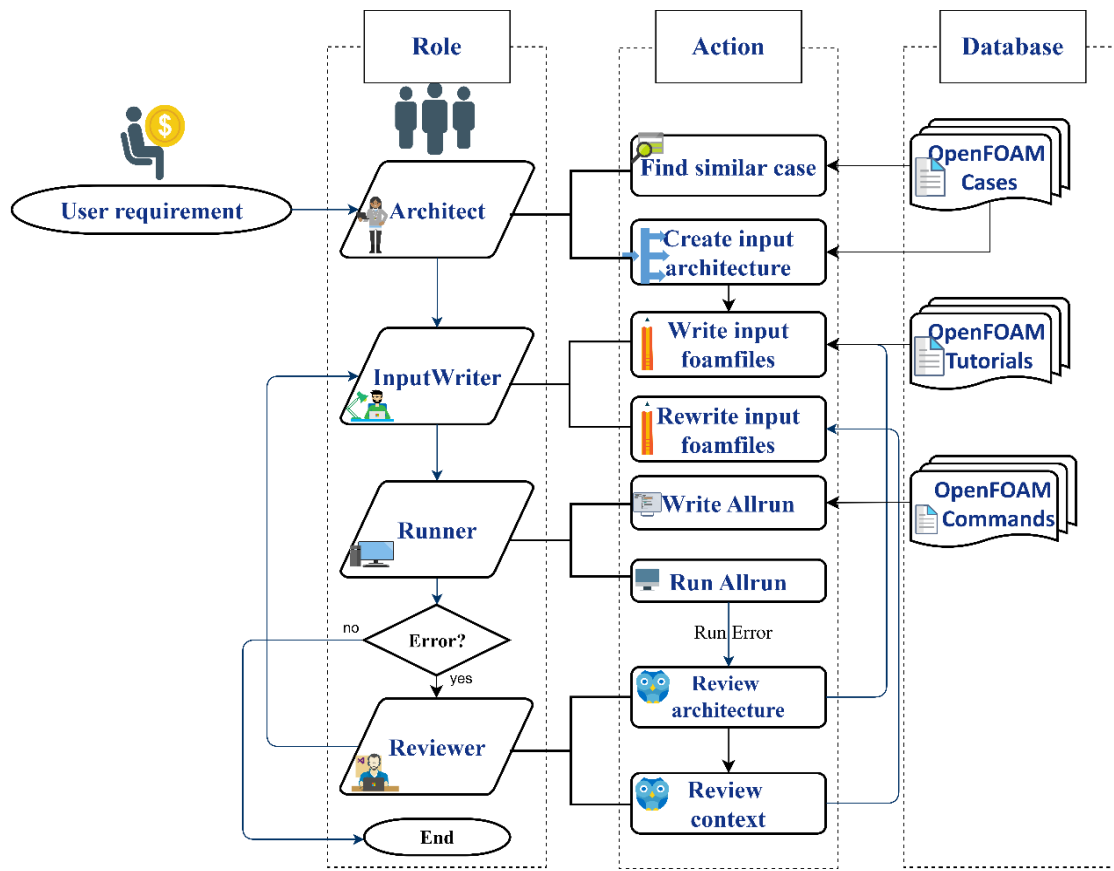

Figure 1. The framework of MetaOpenFOAM

**Runner (Role)**: The Runner executes the CFD simulation using OpenFOAM. This role ensures

that the simulation runs smoothly and monitors for any possible errors during execution.

**Actions of Runner**:

Write the Allrun script to automate the simulation execution process.

Run the Allrun script to perform the CFD simulation.

**Reviewer (Role)**: The Reviewer's role is to analyze any errors that occur during the simulation, identify the relevant file that caused the error, and report these findings back to the InputWriter. This role is to check the file architecture based on existing information to determine whether the error is caused by the absence of related files or the incorrect contents that need modifications.

**Actions of Reviewer**:

Review the input architecture.

Review error context.

**Procedure of MetaOpenFOAM**

First, the Architect takes the user requirements and splits them into several subtasks, which are then given to the InputWriter. The InputWriter creates the necessary OpenFOAM input files and passes them to the Runner. The Runner writes the Allrun script and executes the OpenFOAM simulation. If an error occurs, the Runner provides the execution error command and error context to the Reviewer. The Reviewer examines the file architecture and context to identify and solve the error, then returns the revised instructions to the InputWriter. The InputWriter either writes new files following the revised architecture or corrects the error in the input files. This loop repeats until no errors occur, or it reaches a user-defined maximum number of iterations or maximum investment (based on tokens). The prompts of the main action in each role are shown in Appendix A.

## 2.2 Retrieval-Augmented Generation (RAG) based on Langchain

Langchain's RAG technology is a critical component that supports MetaOpenFOAM by integrating a searchable database of OpenFOAM official documents and tutorials. This system enhances the agents' ability to perform their tasks by providing relevant information and contextual guidance, ensuring that the framework can handle a wide range of CFD scenarios with minimal user intervention.

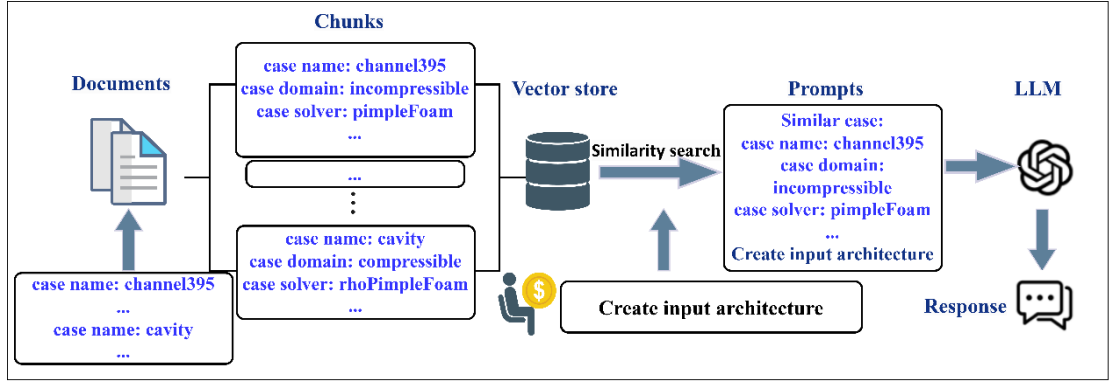

Figure 2. The procedure of Retrieval-Augmented Generation (RAG)

Figure 2 shows the procedure of RAG in action "Find similar case" of MetaOpenFOAM. Firstly, based on the tutorials provided by OpenFOAM, a database containing file structures is constructed. These documents are then split into chunks. Each individual case is divided into separated chunks. A vector store is then created using these chunks. After saving the database, it only requires querying to retrieve the most similar chunks. These chunks are then combined with the user message as input for the LLM, completing the entire RAG process. And for other actions of MetaOpenFOAM, the procedure of RAG is similar, and only the database should be changed into file context or file command, corresponding to the OpenFOAM tutorials and OpenFOAM commands in Figure 1. During the actions "Write input OpenFOAM files" and "Rewrite input OpenFOAM files," documents containing tutorial file contents are required. For the action "Write the Allrun file," documents containing the collection of OpenFOAM execution commands are needed to ensure that the written commands comply with the standards. Specific examples of these documents can be found in Appendix B.

The more detailed document segmentation facilitates more accurate matching during retrieval. The necessity of the above data-enhanced retrieval (RAG) method will be validated in Section 4.1.

## 3. Experiment

### 3.1 Setup

MetaGPT v0.8.0 [11] was selected to integrate various LLMs, while OpenFOAM 10 [2] was utilized for CFD computations due to its stability and reliability as an open-source CFD solver. GPT-4o [15] was chosen as the representative LLM because of its outstanding performance. The temperature, as a parameter of LLM controlling randomness of generated text, was set to 0.01 to ensure highly focused and deterministic outputs, which results in low randomness of the generated text. The influence of temperature on performance is evaluated in Section 4.2.

For RAG technology, LangChain v0.1.19 [21] was employed to link the LLM and the database. The FAISS vector store [22], known for its efficiency and user-friendliness, was used as the database vector store, and OpenAIEmbeddings were selected for embedding the data chunks. The "similarity" method was utilized for matching similar chunks. The combination of retrieved documents and user messages represents the simplest form of stacking. More details could be found in the code: https://github.com/Terry-cyx/MetaOpenFOAM

**3.2   Benchmarking Natural Language Input for CFD Solvers**

Currently, there are no public benchmarks for CFD user requirements to validate CFD solvers that take natural language as input. Therefore, this paper derives several common simulation requirements from OpenFOAM tutorials and lists several relevant simulations needs.

It is important to note that, from the user's perspective, natural language-based CFD user requirements are generally incomplete. They cannot cover all the necessary information required for input files and can only provide the most essential details, such as the case name, case category, solver, mesh, boundary conditions, and initial conditions. Therefore, in the constructed cases below, only partial information is provided as user requirements, aligning with the usage habits of natural language.

Eight cases, covering a range of multidimensional flow problems including both 2D and 3D flows, various compressible and incompressible flows with various physical processes such as turbulence, heat transfer, and combustion, were tested. These cases involve Direct Numerical Simulation (DNS) and the turbulence models of Reynolds-Averaged Navier-Stokes (RANS) and Large Eddy Simulation (LES). Both compressible and incompressible solvers are included, along with description for Newtonian and non-Newtonian fluids. All these cases were modified from OpenFOAM tutorials, with specific parameters adjusted accordingly.

① HIT: do a DNS simulation of incompressible forcing homogeneous isotropic turbulence with Grid 32^3 using dnsFoam
② PitzDaily: do a LES simulation of incompressible pitzDaily flow using pisoFoam with inlet velocity = 5 m/s
③ Cavity: do a 2D RANS simulation of incompressible cavity flow using pisoFoam, with RANS model: RNGkEpsilon, grid 15*15*1
④ LidDrivenCavity: do an incompressible lid driven cavity flow simulation with the top wall moves in the x direction at a speed of 1 m/s while the other three are stationary

⑤ SquareBendLiq: do a compressible simulation of squareBendLiq of using rhoSimpleFoam with endTime = 100, deltaT = 1, and writeInterval = 10
⑥ PlanarPoiseuille: do a laminar simulation of incompressible planar Poiseuille flow of a non-Newtonian fluid with grid 1*20*1, modelled using the Maxwell viscoelastic laminar stress model, initially at rest, constant pressure gradient applied from time zero
⑦ CounterFlowFlame: do a 2D laminar simulation of counterflow flame using reactingFoam in combustion with grid 50*20*1
⑧ BuoyantCavity: do a RANS simulation of buoyantCavity using buoyantFoam, which investigates natural convection in a heat cavity with a temperature difference of 20K is maintained between the hot and cold; the remaining patches are treated as adiabatic.

**3.3 Evaluation Metrics**

It is essential to establish evaluation metrics to assess the performance of natural language-based CFD solvers. These metrics need to consider common indicators in the CFD domain, such as successful mesh generation and convergence of calculations, as well as computational costs and generation success rates (pass@k in [23]) relevant to the LLM domain. Therefore, we evaluate the practical performance of natural language-based CFD solvers using the following five metrics: the first four metrics A, B, C, D for single experiments and the last metric E being added for multiple experiments.

Single experiments can be evaluated by the following metrics:

**(A) Executability**: This metric rates input files on a scale from 0 (failure) to 4 (flawless). A score of '0' indicates grid generation failure, '1' indicates grid generation success but running failure, '2' indicates that the case is runnable but does not converge, '3' indicates that the case runs to the endTime specified in the controlDict, and '4' indicates flawless input foam files, which not only run to the endTime but also meet all user requirements. A single experiment is considered to have passed the test if its executability score reaches '4'. Among them, '1' to '3' can be judged automatically by the program, but '4' requires human participation.

**(B) Cost**: The cost evaluation includes (1) running time, (2) number of iterations, (3) token usage, and (4) expenses, which are proportional to token usage.

**(C) Code Statistics**: This metric includes (1) the number of input files, (2) the number of lines per input file, and (3) the total number of lines in the input files.

**(D) Productivity**: This metric is defined as the number of tokens used divided by the number of lines in the input files, representing token consumption per input line.

For multiple experiments, a new metric needs to be added:

**(E) Pass@k** [23]: This metric represents the probability that at least one of the k generated input file samples passes the unit tests. It measures the model's ability to generate correct input files within k attempts. We follow the unbiased version of pass@k as presented by Chen et al. (2021a) and Dong et al. (2023) to evaluate the pass@k of MetaOpenFOAM

$$pass@k := \underset{problems}{E}\left[1 - \frac{\binom{n-c}{k}}{\binom{n}{k}}\right],$$

where $n$ represents the number of input sets generated for each user requirement, and $c$ represents the number of these samples that pass the test, i.e., achieve an executability score of 4. To evaluate pass@k, we generate $n \geqslant k$ samples per task (with $n = 10$ and $k = 1$ in this paper), count the number of correct samples $c \leqslant n$ which pass unit tests, and calculate the unbiased estimator.

### 3.4 Main results

Table 1 Performance of MetaOpenFOAM

|  | Executability | Token Usage | Iteration | Productivity | Pass@1(%) |
|---|---|---|---|---|---|
| **HIT** | 4 | 12667 | 2.4 | 36.3 | 100 |
| **PitzDaily** | 4 | 18083 | 2.1 | 32.1 | 100 |
| **Cavity** | 4 | 12863 | 0 | 28.3 | 100 |
| **LidDrivenCavity** | 2.8 | 52090 | 12.5 | 149.7 | 60 |
| **SquareBendLiq** | 4 | 16385 | 0 | 27.6 | 100 |
| **PlanarPoiseuille** | 3.7 | 35532 | 5.2 | 81.3 | 90 |
| **CounterFlowFlame** | 3.7 | 47927 | 7.2 | 20.3 | 90 |
| **BuoyantCavity** | 2.4 | 156812 | 16.3 | 161.3 | 40 |
| **Average** | 3.6 | 44045 | 5.7 | 67.1 | 85 |

Table 1 presents the performance of MetaOpenFOAM on eight test cases from the benchmark proposed in Section 3.2. The complete flow chart for the individual examples is described in detail in Appendix C, using HIT as an example. Only a selection of key metrics from Section 3.3 is displayed, while the rest are provided in Appendix D. Each metric for single experiments is the average of n tests (n=10). For the iteration metric in the cost, maximum iteration is set to 20 in the program to prevent infinite iterations. If executability does not reach 3 or above after 20 iterations, the test is automatically marked as failed and the iteration is terminated.

Overall, the average pass rate (pass@1) of 85% and high executability score 3.6 demonstrate the outstanding performance of MetaOpenFOAM. On average, each test case requires 44,045 tokens. Given a cost of $5 per 1M tokens, generating one test case costs only $0.22, and each line of input file consumes an average of 67.1 tokens, costing just $0.0003, which is significantly lower than manual labor costs. Therefore, in terms of lowering the barrier to use, reducing labor costs, and increasing efficiency, MetaOpenFOAM is revolutionary.

For different test cases, it can be seen that HIT, PitzDaily, Cavity, and SquareBendLiq have an executability score of 4, meaning flawless results that satisfy all user requirements. These cases require fewer iterations and fewer tokens. However, for cases with lower executability scores, such as BuoyantCavity and LidDrivenCavity, the LLM fails to correctly modify errors, resulting in more iterations and higher token usage, and consequently, a lower pass@1.

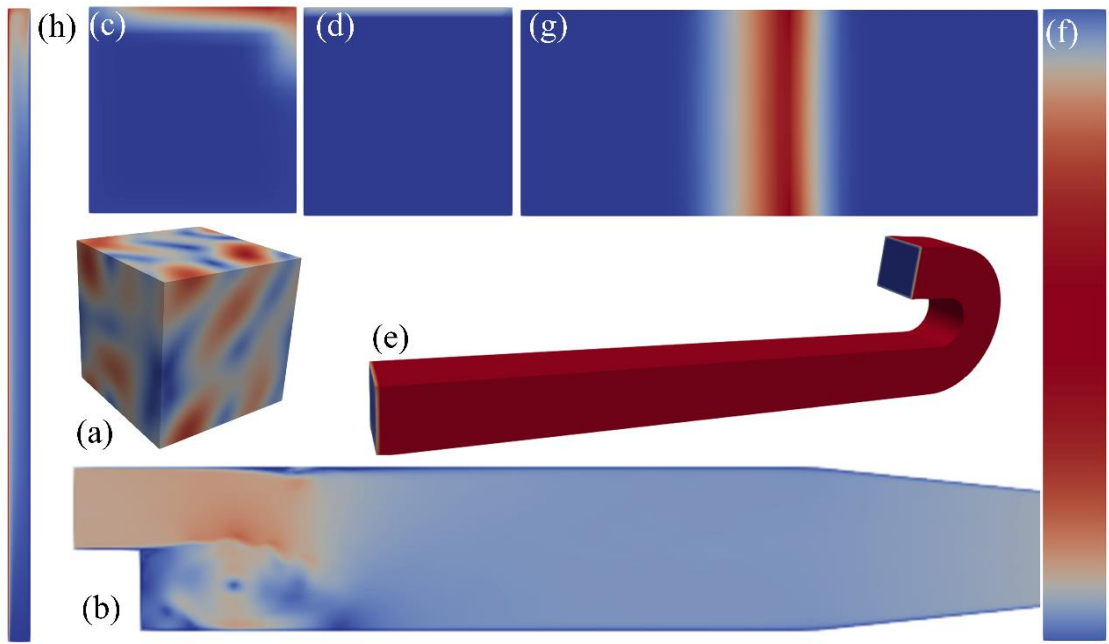

Figure 3. Demo simulation results simulated by MetaOpenFOAM. (a) Homogeneous Isotropic Turbulence (HIT) (b) PitzDaily (c) Cavity (d) Lid-Driven Cavity (e) Square Bend Liquid (SquareBendLiq) (f) Planar Poiseuille (g) Counter Flow Flame (h) Buoyant Cavity.

Analyzing the correlation between iteration and token usage, we find a Pearson correlation coefficient of 0.89 with a p-value of 0.0013. This indicates a strong positive correlation between iteration and token usage, and this correlation is statistically significant (p-value much less than 0.05). The reason is evident: more iterations mean more input to the LLM. Therefore, in the subsequent discussion, we will focus only on iteration and not token usage, due to their strong correlation.

As shown in Figure 3, we visualized MetaOpenFOAM's contour plots for eight test examples using paraview [24]. Detailed results about performance and test examples are provided in Appendix D.

**4. Discussion**

This section will discuss the necessity of each component in MetaOpenFOAM as well as sensitivity analysis of key parameters related to large language models and qualitative analysis of some MetaOpenFOAM results.

**4.1 Ablation Analysis**

To validate the necessity of each component in MetaOpenFOAM, we removed some components to quantitatively discuss how their removal influences the model outputs. Specifically, since MetaOpenFOAM primarily consists of Roles, Actions, and RAG, the ablation study focused on these three aspects.

A. Remove "Reviewer" Role

As shown in section 2.1, there are 4 roles in MetaOpenFOAM where the Architect, InputWriter, and Runner are essential roles. Removing any of these would result in an executability of 0, making ablation analysis meaningless. However, the Reviewer is not strictly necessary, allowing for comparison with and without it. It is important to note that the Reviewer is the key component of MetaOpenFOAM that distinguishes automated multi-agent collaboration from a simple aggregation of single agents. Therefore, examining how the Reviewer affects model outputs is vital.

B. Remove "Review architecture" Action

Each role involves several actions. Some actions, like creating the input architecture, writing OpenFOAM input files, and running OpenFOAM, are essential. Omitting these actions would obviously prevent the tests from passing. Therefore, we only considered the non-essential action of "Review architecture". If "Review architecture" is missing, errors related to wrong file architecture become difficult to resolve.

C. Remove RAG

RAG technology is another key component of MetaOpenFOAM, providing professional knowledge missing from general LLMs. To verify the importance of RAG, we tested the performance of MetaOpenFOAM without it.

Table 2: The ablation study on role, action and RAG. '#' denotes 'Remove', '#Reviewer' means

remove Reviewer, '#Nothing' means remove nothing, i.e., the complete and original MetaOpenFOAM, 'Arc.' denotes the input architecture.

| Statistical Index | #Reviewer | #Review Arc. | #RAG | #Nothing |
|---|---|---|---|---|
| (A) Executability | 1.7 | 3.0 | 0.8 | **3.6** |
| (B) Cost: Running time (s) | **126** | 292 | 332 | 271 |
| (B) Cost: Iteration (max: 20) | **0** | 7.7 | 19.2 | 5.7 |
| (B) Cost: Token Usage | **15661** | 69336 | 81346 | 44045 |
| (C) Code Statistic: Input Files | 12.8 | 13.2 | **10.1** | 13.7 |
| (C) Code Statistic: Lines per Input File | **42.0** | 42.5 | 124.5 | 49.9 |
| (C) Code Statistic: Total lines of Input Files | **540** | 572 | 1491 | 760 |
| (D) Productivity | 29.8 | 143.0 | 158.2 | 67.1 |
| (E) pass@1 (%) | 27.5 | 70 | 0 | **85** |

Table 2 shows the performance of MetaOpenFOAM when the Reviewer (role), Review input architecture (action), RAG, or nothing is removed. After removing the Reviewer role, the pass@1 of MetaOpenFOAM dropped from 85% to 27.5%. Without the Reviewer, no iterations occurred, resulting in lower running time and cost. However, the significantly lower pass@1 and executability indicate a substantial decline in MetaOpenFOAM's applicability and utility. As shown in Figure 4, apart from the cavity and squareBendLiq cases, which require no iterations, all other cases failed. This demonstrates that without the Reviewer role, MetaOpenFOAM cannot handle CFD simulation tasks with moderate complexity. When the "Review architecture" action was removed, the pass@1 also decreased from 85% to 70%. This decline mainly stemmed from the HIT and LidDrivenCavity cases, where file architecture-related errors (for example, "cannot find file XX") occurred. Without the "Review architecture" action, MetaOpenFOAM could not create a new file architecture to resolve these issues, resulting in repeated modifications to existing files. Thus, the "Review architecture" action is verified as necessary.

For the RAG module, another key component of MetaOpenFOAM, removing RAG resulted in a pass@1 of 0, indicating that MetaOpenFOAM without RAG could not pass any CFD simulation tasks in the database. However, executability was not 0 because some simulation tasks could complete mesh generation (Executability=1), run but not converge (Executability=2), or run successfully but not meet user requirements (Executability=3). For instance, in the PitzDaily test, MetaOpenFOAM without RAG could even achieve an executability of 3, but the simulated case used a randomly generated 20*20*20 cubic mesh instead of the PitzDaily mesh. This shows that without RAG, MetaOpenFOAM loses the ability to complete CFD simulation tasks, likely due to the lack of training data for constructing OpenFOAM input files in LLM.

Detailed data for the entire ablation study are presented in table form in Appendix E.

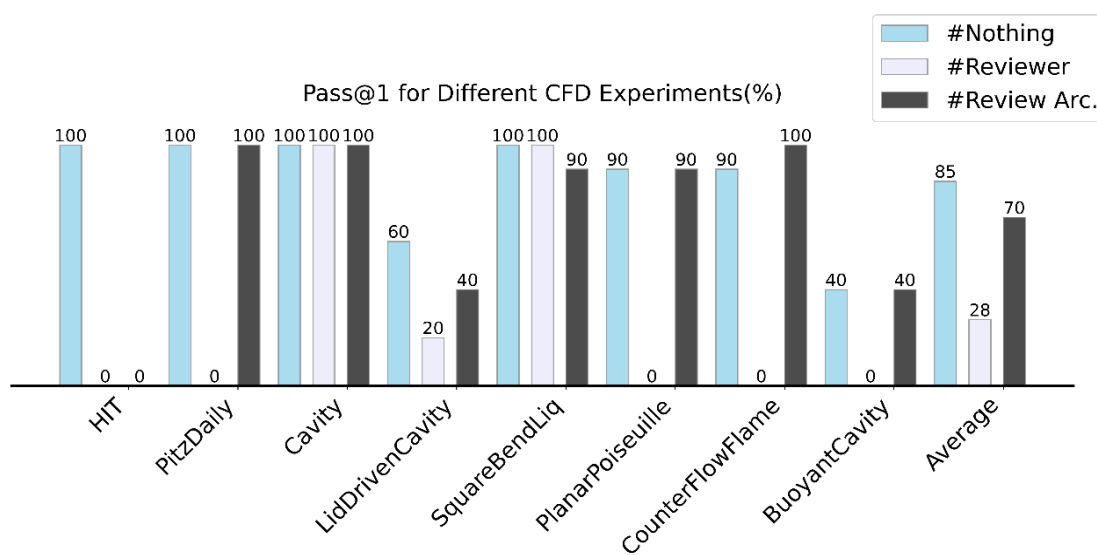

Figure 4 pass@1 of ablation study on role, action. '#' denotes 'Remove', '#Reviewer' means remove Reviewer, '#Nothing' means remove nothing, i.e., the complete and original MetaOpenFOAM, 'Arc.' denotes the input architecture.

### 4.2 Influence of parameter "temperature"

In LLM, the "temperature" is a key parameter, which controls the randomness and creativity of the generated text. When the temperature is high (for example near 1), the probability distribution becomes flatter, making the model more likely to choose words with less probability. The generated text becomes more diverse and creative, but it may also be less coherent or sensible. When the temperature is low (for example near 0), the probability distribution becomes sharper, making the model more likely to choose the most probable words. The generated text becomes more conservative and coherent but less diverse and creative.

Table 3. The sensitivity analyzes of parameter 'temperature' of LLM, where 'temp.' means

'temperature'.

| Statistical Index | temp.=0.01 | temp.=0.5 | temp.=0.99 |
|---|---|---|---|
| (A) Executability | **3.6** | 3.4 | 2.3 |
| (B) Cost: Running time (s) | **271** | 357 | 350 |
| (B) Cost: Iiterations (max: 20) | **5.7** | 7.4 | 13.5 |
| (B) Cost: Token Usage | **44045** | 58419 | 109195 |
| (C) Code Statistic: Input Files | **13.7** | 14.4 | 14.1 |
| (C) Code Statistic: Lines per Input File | 49.9 | 59.9 | **43.3** |
| (C) Code Statistic: Total lines of Input Files | 760 | 874 | **614** |
| (D) Productivity | **67.1** | 87.5 | 179.3 |
| (E) pass@1 (%) | **85** | 83 | 48 |

Table 3 shows the average evaluation metrics after running the database of test cases 10 times at three different temperatures (0.01, 0.5, 0.99). It is evident that when the temperature is set to 0.01, both (A) Executability and (B) Cost, as well as (D) Productivity and (E) pass@1, are optimal on average. This suggests that conservative and coherent generated text leads to more stable and accurate results.

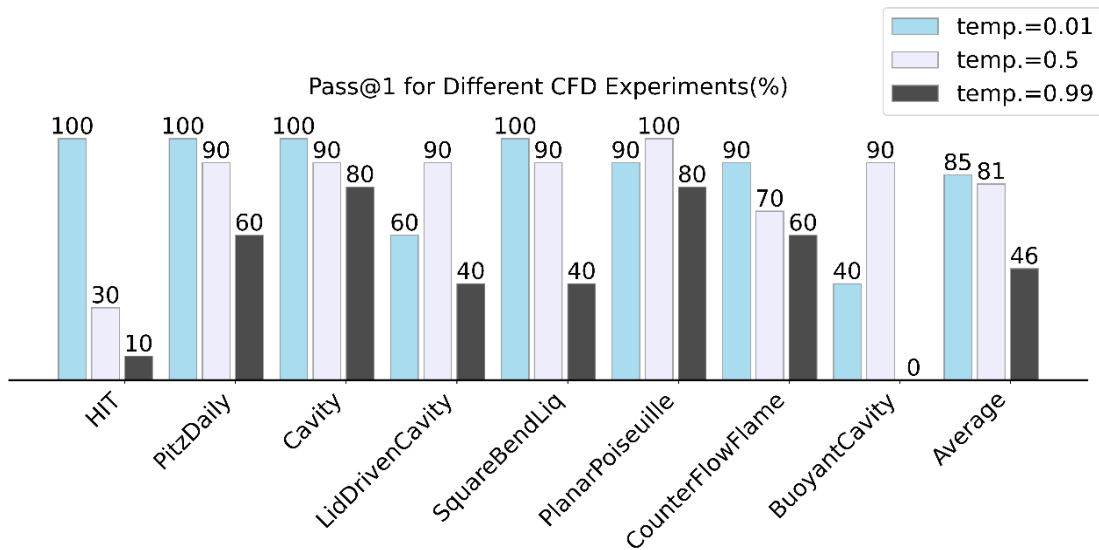

Figure 5 Average pass@1 for different CFD experiments with 3 different temperatures (0,01, 0.5, 0.99) of LLM, where 'temp.' means 'temperature'.

Figure 5 shows the average pass@1 for the eight test cases under three different temperatures

(0.01, 0.5, 0.99). We found that for most case, lower temperature leads to higher pass@1 due to the more stable and accurate generated inputs. However, for a few cases, such as LidDrivenCavity and BuoyantCavity, a middle temperature (temperature=0.5) achieved better pass@1 than the low temperature (temperature=0.01). But high temperature (temperature = 0.99) results in low pass@1 in all the test dataset, which means that LLM with high temperature is not suitable for CFD tasks solving.

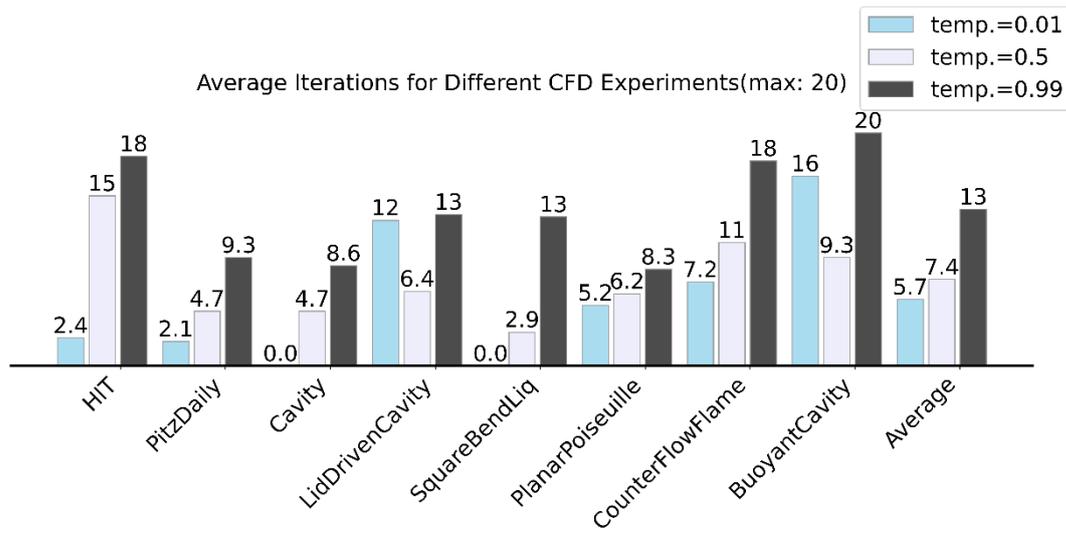

Figure 6 Average iterations for different CFD experiments with 3 different temperatures of LLM

Figure 6 illustrates the average iterations for eight test cases under three different temperatures (0.01, 0.5, 0.99). Each iteration requires running the InputWriter, Runner, and Reviewer. Iterations is a cost metric for MetaOpenFOAM, which correlates positively with tokens and expenses. For most cases, lower temperature leads to lower iterations due to the more stable and accurate generated inputs. However, for a few cases like LidDrivenCavity and BuoyantCavity, very low randomness (temperature=0.01) led to overly sharp probability distributions, resulting in repeated errors during iterations. In contrast, moderate randomness (temperature=0.5) helped overcome the fixed patterns of errors produced at low temperatures, reducing the number of iterations and thereby lowering tokens and expenses.

Therefore, a dynamic temperature model might be beneficial. It could dynamically adjust from low to middle temperatures when low-temperature generation is ineffective, improving pass rates and reducing costs in MetaOpenFOAM.

## 4.3 Generalizability Study

In this section, we will qualitatively analyze the performance of MetaOpenFOAM in some special situations such as the ones requiring modification of key parameters, matching with less similar cases, or requiring human participation to improve performance.

A. Key data identification and modification

One of the most important capabilities in a natural language based CFD simulation framework is the ability to identify changes to key data in the user requirement and modify those changes when writing the input file. This capability is the foundation of CFD numerical optimization based on natural language and is a key differentiator from executability 3 to 4 in evaluation metrics.

Key data can be divided into two categories. One is can be directly corrected in the input, such as grid size, model parameters, and control parameters; the other category cannot be directly modified in the input, such as the Reynolds number.

For key data that can be directly corrected in the input, MetaOpenFOAM can identify and modify the corresponding parameters in the input. Here lists the dataset where the key data changed:

① HIT: do a DNS simulation of incompressible forcing homogeneous isotropic turbulence with Grid 20^3 using dnsFoam. [32->20]
② PitzDaily: do a LES simulation of incompressible pitzDaily flow using pisoFoam with inlet velocity = 8 m/s. [5->8]
③ Cavity: do a 2D RANS simulation of incompressible cavity flow using pisoFoam, with RANS model: KEpsilon, grid 15*15*1. [RNGkEpsilon-> KEpsilon]
④ LidDrivenCavity: do an incompressible lid driven cavity flow simulation with the top wall moves in the x direction at a speed of 2 m/s while the other 3 are stationary. [1->2]
⑤ SquareBendLiq: do a compressible simulation of squareBendLiq of using rhoSimpleFoam with endTime = 1000, deltaT = 1, and writeInterval = 100. [100->1000, 10->100]
⑥ PlanarPoiseuille: do a laminar simulation of incompressible planar Poiseuille flow of a Newtonian fluid with grid 1*20*1, initially at rest, constant pressure gradient applied from time zero. [non-Newtonian -> Newtonian]
⑦ CounterFlowFlame: do a 2D laminar simulation of counterflow flame using reactingFoam in combustion with grid 50*20*1. [50*20*1->40*20*1]
⑧ BuoyantCavity: do a RANS simulation of buoyantCavity using buoyantFoam, which investigates natural convection in a heat cavity with a temperature difference of 15K is maintained between the hot and cold; the remaining patches are treated as adiabatic. [20->15]

Table 4 presents the performance test results of MetaOpenFOAM on the original dataset (dataset1) and the modified dataset (dataset2). It was found that the two datasets show similar performance in terms of Executability and pass@1, indicating that MetaOpenFOAM can identify and modify key parameters in the input for key data that can be directly modified, achieving similar performance as the original dataset.

Figure 7 displays the executability of each case before and after modification. The differences between dataset1 and dataset2 are minimal for most cases, except for SquareBendLiq, which shows a significant difference. This discrepancy arises because the user requirements for SquareBendLiq demand a longer simulation time. During this extended simulation time, the case encountered convergence issues, resulting in an Executability of 2 and a pass@1 of 0.

Table 4. Performance of MetaOpenFOAM in two different datasets (dataset1: original dataset, dataset2: modified dataset)

| Statistical Index | Dataset1 | Dataset2 |
| --- | --- | --- |
| (A) Executability | 3.6 | 3.7 |
| (B) Cost: Running time (s) | 271 | 294 |
| (B) Cost: Iteration (max: 20) | 5.7 | 7.2 |
| (B) Cost: Token Usage | 44045 | 48416 |
| (C) Code Statistic: Input Files | 13.7 | 13.5 |
| (C) Code Statistic: Lines per Input File | 50 | 42 |
| (C) Code Statistic: Total lines of Input Files | 760 | 581 |
| (D) Productivity | 67.1 | 79.2 |
| (E) pass@1 (%) | 85 | 85 |

However, for some data that cannot be directly modified in the input, such as Reynolds number, the current version of MetaOpenFOAM cannot directly recognize such data that require secondary calculation.

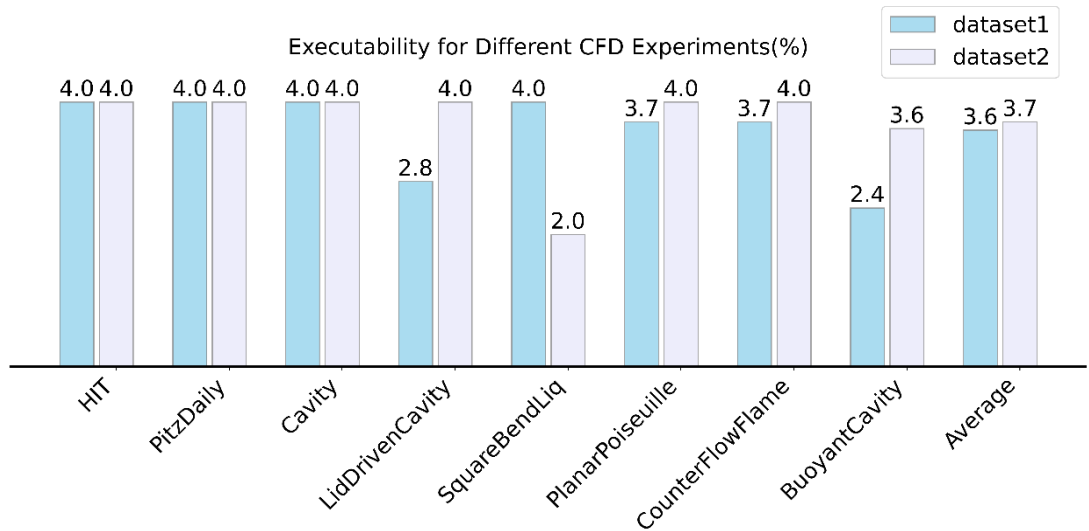

Figure 7 Average executability for different CFD experiments in two different datasets (dataset1: original dataset, dataset2: modified dataset)

For example, in the HIT case, if the user requests a simulation with Reynolds number at 20, MetaOpenFOAM cannot produce a simulation that meets user requirement. This issue is not insurmountable. In the future, by introducing additional agents and a comprehensive library of functions to convert key parameters to input parameters, MetaOpenFOAM could handle a broader range of CFD simulation tasks.

B. Failure similarity match

Similarity match, as a part of RAG technology is performed during the "Find similar cases" action by the Architect. However, it may fail to match the most similar case due to the inaccurate user requirement, or the inadequate database, which may not contain similar cases. In such scenarios, both pass@1 and executability are significantly reduced. However, even in these cases, using RAG technology still yields better results than not using it.

For example, in the lidDrivenCavity case, the most similar case found in 10 tests was sloshingCylinder, which clearly does not have a high similarity match. Despite this, shown as in Table 1 and Table 8, the pass@1 60% and executability 2.8 with RAG technology were still much better than without it, whose pass@1 is 0% and executability is 1.0.

C. Human participation

While MetaOpenFOAM aims to perform successful simulations using natural language inputs alone, for cases with complex geometries and boundary conditions, natural language inputs alone are insufficient. Therefore, human participation is needed to enhance the performance of

MetaOpenFOAM. Interfaces have been included in MetaOpenFOAM to handle issues related to complex geometries and boundary conditions.

These interfaces allow for human intervention to refine the simulation setup, ensuring that MetaOpenFOAM can tackle more sophisticated CFD tasks that are beyond the current capabilities of fully automated processes. This combination of automated and human-assisted approaches aims to strike a balance between efficiency and accuracy in CFD simulations.

5. **Conclusion**

For the first time, an LLM-based multi-agent framework for CFD has been established, incorporating the construction of a multi-agent system based on MetaGPT and RAG technology based on Langchain. Specifically, the multi-agent system consists of four roles: Architect, InputWriter, Runner, and Reviewer. These roles work together by first decomposing the CFD simulation task into a series of input file generation subtasks (Architect), then generating the corresponding input files based on the subtasks (InputWriter), running the CFD simulation (Runner), and finally providing feedback to the InputWriter by checking the simulation results (Reviewer) until the results meet expectations or the maximum number of iterations is reached. RAG technology aids this process by converting OpenFOAM tutorials into a database, matching the most similar cases from the database whenever the LLM needs access, thereby assisting the LLM in answering queries.

To test the performance of MetaOpenFOAM in CFD simulations, a benchmark for natural language-based CFD solvers was proposed, consisting of eight CFD simulation tasks derived from OpenFOAM tutorials. Tests on the benchmark have shown that MetaOpenFOAM achieved a high pass@1 rate of 85%, with each test case costing only \$0.22 on average. This demonstrates MetaOpenFOAM's ability to automate CFD simulations using only natural language input and iteratively correct errors to achieve the desired simulation at a low cost.

An ablation study was conducted to verify the roles of each component in the multi-agent system and the RAG technology. The results showed that removing the Reviewer, Action review architecture, and RAG resulted in pass@1 rates of 27.5%, 70%, and 0%, respectively, all lower than the original framework's 85%. This demonstrates the necessity of reviewing results and RAG technology. A sensitivity study on the temperature parameter in the LLM showed that compared to middle/high temperature (0.5/0.99) pass@1 rates of 83% and 48%, the LLM based on low temperature (0.01) has a higher pass@1 rate of 85% on average. This suggests that conservative and coherent generated text leads to more stable and accurate results. However, the middle temperature (0.5) also showed good performance in specific cases.

Additionally, MetaOpenFOAM maintained an 85% pass@1 rate even after modifying keywords in the prompts, indicating its ability to identify and modify key parameters in the input corresponding to the keywords. The performance on failure similarity matches also showed that MetaOpenFOAM has good simulation capability even when the most similar case is not matched. Finally, incorporating human participation can enhance the performance of MetaOpenFOAM in handling issues related to complex geometries and boundary conditions. These three aspects demonstrate that MetaOpenFOAM possesses a certain degree of generalization ability.

**Acknowledgments**

This work was supported by the National Natural Science Foundation of China (No. 52025062 and 52106166). The authors also acknowledge High-Performance Computing Centre at Tsinghua University for providing computational resource.

During the preparation of this work the author(s) used ChatGPT in order to improve language and readability. After using this tool/service, the author(s) reviewed and edited the content as needed and take(s) full responsibility for the content of the publication.

**Appendix A. Prompts of MetaOpenFOAM**

In this Appendix, we will present some key action prompts for the four roles: Architect,

InputWriter, Runner, and Reviewer. For the complete prompts, please refer directly to the source code: https://github.com/Terry-cyx/MetaOpenFOAM

Action: Create input architecture in Role: Architect

```
    User requirement:
    {requirement}
    Your task is to generate the OpenFOAM input foamfiles list following file structure of OpenFOAM cases to meet the user requirements.
    Here is a OpenFOAM case similar to the user requirements
    The following is a case of OpenFOAM, which is similar to the user's requirements:
    {tutorial}
    Please take this case as a reference. generate the OpenFOAM input foamfiles list following file structure of OpenFOAM cases to meet the user requirements.
    You should splits foamfiles list into several subtasks, and one subtask corresponds to one input foamfile
    Return ```splits into number_of_subtasks subtasks:
    subtask1: to Write a OpenFoam specific_file_name foamfile in specific_folder_name folder that could be used to meet user requirement:{requirement}.
    subtask2: to Write a OpenFoam specific_file_name foamfile in specific_folder_name folder that could be used to meet user requirement:{requirement}.
    ...

    ``` with NO other texts,
    your subtasks:
```

Action: Write input file in Role: InputWriter

```
    Your task is {requirement}.
    The similar foamfile is provided as follows:
    {tutorial_file}
    Please take this foamfile as a reference, which may help you to finish your task.
    According to your task, return ```your_code_here``` with NO other texts,
    your code:
```

Action: Write allrun file in Role: Runner

```
        Your task is to write linux execution command allrun file to meet the user requirement: {requirement}.
        The input file list is {file_list}.
        Here is a OpenFOAM allrun file similar to the user requirements:
```

```
        {tutorial}
        Please take this file as a reference.
        The possible command list is
        {commands}
        The possible run list is
        {runlists}
        Make sure the written linux execution command are coming from the
above two lists.
        According to your task, return ```your_allrun_file_here``` with NO
other texts
```

Action: Review file architecture in Role: Reviewer

```
    {command} has been executed in OpenFOAM10, and got the following error:
    {error}
    The corresponding input file list is:
    {file_list} in folder {folder_list}
    Please analyze which files the error may be related to, and return the
related files and the corresponding folders as follows:
    ###file_name1, file_name2, ...### in ```file_folder1,
file_folder2, ...``` with NO other texts.
    where file_name1, file_name2, ..., come from {file_list}
    and file_folder1, file_folder2, ..., come from {folder_list}
```

Action: Review file context in Role: Reviewer

```
    to rewrite a OpenFoam {file_name} foamfile in {file_folder} folder that
could solve the error:
    ###ERROR BEGIN:
    {error}
    ERROR END.###
    Note that {file_list} in {folder_list} folder was found to be
associated with the error, and you need to rewrite {file_name} first,
taking into account how these files affect each other.
    the original {file_list} in {folder_list} folder encounter the error
when {command} has been executed in OpenFOAM10,
    {related_files}
    Note that you need to return the entire modified file, never return a
single modified fragment, because I want to save and run the file directly,
making sure there are no other characters
    According to your task, return ```your_code_here``` with NO other
texts,
    your code:
```

## Appendix B. OpenFOAM database

The OpenFOAM database is derived from the tutorials in the main OpenFOAM directory and

is primarily divided into three sub-databases: foamfile architecture, foamfile context, and allrun files. Different sub-databases are used based on the specific action being performed. Below are some excerpts:

Sub-database foamfile architecture:

```
###case begin:
case name: pitzDaily
case domain: compressible
case category: LES
case solver: rhoPimpleFoam
case input name:['fvSolution', 'fvSchemes', 'fvConstraints', 'controlDict',
'momentumTransport', 'physicalProperties', 'nut', 'k', 'T', 'p', 'alphat',
'U', 'muTilda']
corresponding input folder:{'fvSolution': 'system', 'fvSchemes': 'system',
'fvConstraints': 'system', 'controlDict': 'system', 'momentumTransport':
'constant', 'physicalProperties': 'constant', 'nut': '0', 'k': '0', 'T':
'0', 'p': '0', 'alphat': '0', 'U': '0', 'muTilda': '0'}
case end.###
```

Sub-database foamfile context:

```
/*--------------------------------*- C++ -*----------------------------------*\
  =========                 |
  \\      /  F ield         | OpenFOAM: The Open Source CFD Toolbox
   \\    /   O peration     | Website:  https://OpenFOAM.org
    \\  /    A nd           | Version:  10
     \\/     M anipulation  |
\*---------------------------------------------------------------------------*/
FoamFile
{
    format      ascii;
    class       dictionary;
    object      fvConstraints;
}
// * * * * * * * * * * * * * * * * * * * * * * * * * * * * * * * * * * * * * //
limitp
{
    type        limitPressure;
    minFactor   0.5;
    maxFactor   2;
}
```

```
//
*************************************************************************
//
input_file_end.```
```

Sub-database allrun files:

```
```input_file_begin: linux execution command allrun file of case pitzDaily
(domain: compressible, category: LES, solver:rhoPimpleFoam):
#!/bin/sh
cd ${0%/*} || exit 1    # Run from this directory

# Source tutorial run functions
. $WM_PROJECT_DIR/bin/tools/RunFunctions

application="$(getApplication)"

runApplication blockMesh -dict
$FOAM_TUTORIALS/resources/blockMesh/pitzDaily
runApplication $application

#--------------------------------------------------------------------------
----
input_file_end.```
```

**Appendix C. DNS HIT complete flowchart**

In this section, we use the HIT (Homogeneous Isotropic Turbulence) example to demonstrate the basic workflow of MetaOpenFOAM. The corresponding user requirement is to "do a DNS simulation of incompressible forcing homogeneous isotropic turbulence with Grid $32^3$ using dnsFoam."

In this user requirement, we provide the case description, category, solver, model, and grid information, while omitting details such as boundary conditions, initial conditions, run time, and time steps. This approach aligns with typical user practices.

First, based on the user requirement, the Architect converts it into a standardized format, including case name, case domain, case category, and case solver. Then, it searches the database for the most similar case, retrieves its file architecture, and writes a new file architecture suitable for the current case based on the similar case's architecture and user requirements. This new file architecture is divided into multiple subtasks and passed to the InputWriter. It is important to note that each file to be generated in the architecture corresponds to a subtask.

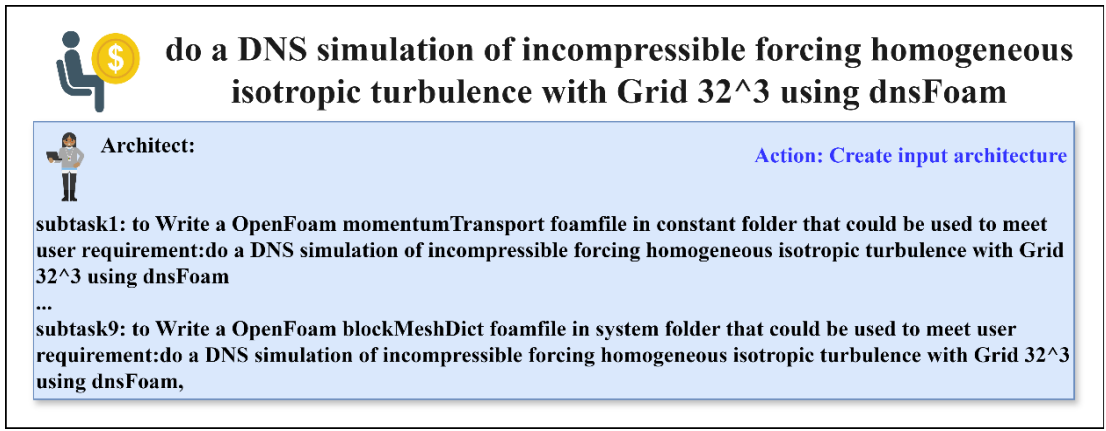

Figure 8. Sketch of Action Create input architecture in HIT case

Next, the InputWriter uses the information provided by the Architect, along with the file information from the similar case in the database, to write the input files for each subtask. Once all subtasks are completed, the work is passed to the Runner. The Runner writes an Allrun file suitable for the current requirement based on the Allrun file information from the similar case in the database and executes it in the OpenFOAM environment. If there are no errors and the Executability >= 3, the loop exits. However, when executing the HIT example, an error "No 'neighbourPatch' provided" was encountered. Therefore, the Runner passes this error information to the Reviewer for correction.

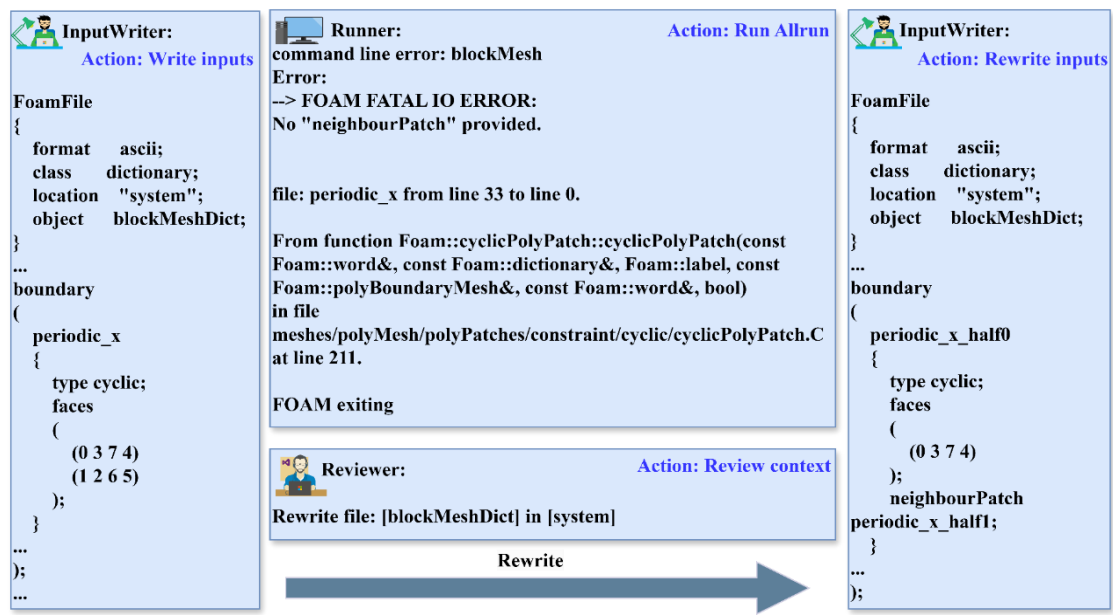

Figure 9. Sketch of the first iteration using MetaOpenFOAM in HIT case

The Reviewer first determines whether the error is related to the file architecture or the file content. In the HIT example, the Reviewer identifies the error as being related to the content of the blockMeshDict file, so it is returned to the InputWriter for modification. This entire process constitutes the first iteration.

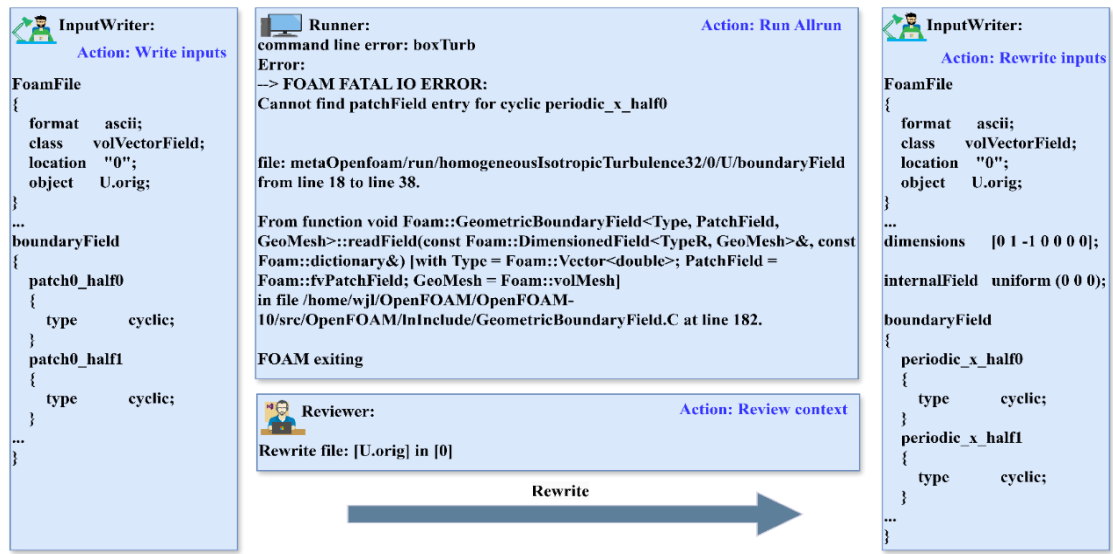

Figure 10. Sketch of the second iteration using MetaOpenFOAM in HIT case

The Runner then continues to execute the modified input files but encounters a new error: "cannot find patchField entry for cyclic periodic_x_half0," and passes this to the Reviewer. The Reviewer determines that the issue lies with the content of the U.orig file, so it instructs the InputWriter to rewrite U.orig. The Runner then runs the simulation again, ultimately achieving successful execution of the HIT example with Executability = 4.

**Appendix D. Detailed performance and test examples of main results**

Table 5. Performance of MetaOpenFOAM

|  | Running time (s) | Prompt Tokens | Completion Tokens | Number of input files | Lines per input file | Total lines of input files |
|---|---|---|---|---|---|---|
| **HIT** | 189 | 9181 | 3486.4 | 10.8 | 32.3 | 348.7 |
| **PitzDaily** | 649 | 13426 | 4656.8 | 11 | 51.2 | 563.7 |
| **Cavity** | 98 | 8709 | 4154.5 | 12 | 37.8 | 454 |
| **LidDrivenCavity** | 200 | 45253 | 6837.5 | 9.2 | 38 | 347.9 |
| **SquareBendLiq** | 154 | 10562 | 5823.4 | 15 | 39.5 | 593 |
| **PlanarPoiseuille** | 181 | 30122 | 5410 | 11 | 40 | 437 |
| **CounterFlowFlame** | 260 | 37459 | 10469.2 | 20.4 | 111 | 2363.4 |
| **BuoyantCavity** | 439 | 142209 | 14602.8 | 19.8 | 49.01 | 972 |
| **Average** | 418.375 | 45906 | 7774 | 13.65 | 50 | 760 |

Where "Prompt Tokens" refer to the tokens required for the prompts sent to the LLM, while "Completion Tokens" are the tokens needed for the LLM's responses. In MetaOpenFOAM, Prompt Tokens are often more numerous than Completion Tokens. This trend becomes more pronounced with an increasing number of iterations. This is because identifying the cause of errors necessitates transmitting all input files to the LLM. The more iterations there are, the more frequently the input files are transmitted, resulting in a higher number of Prompt Tokens compared to Completion Tokens.

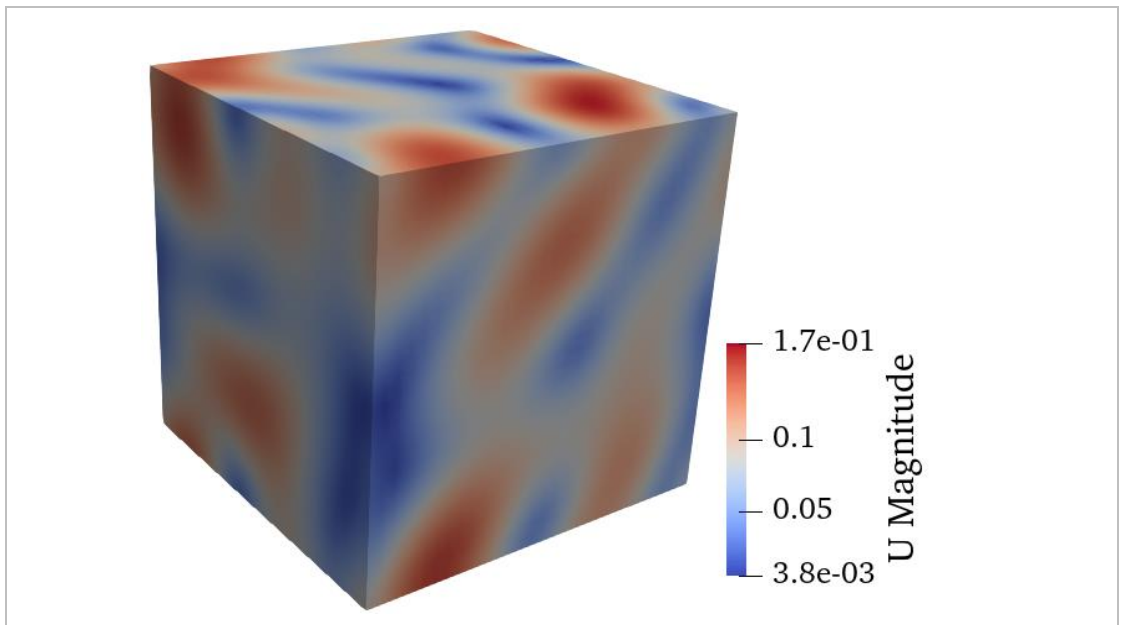

(a)

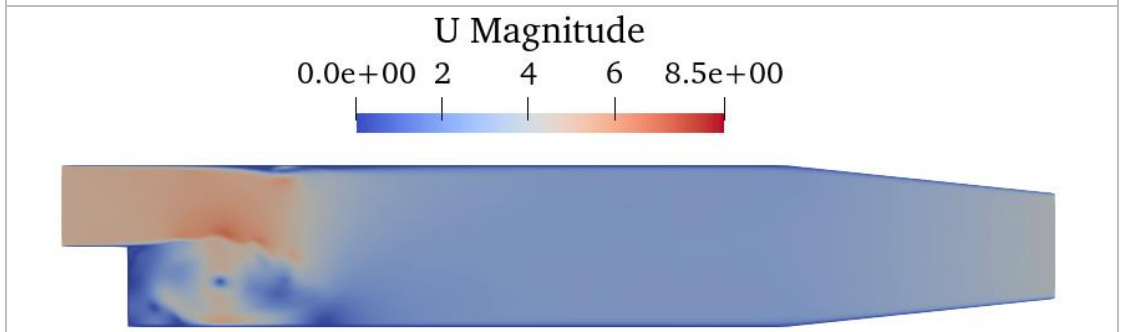

(b)

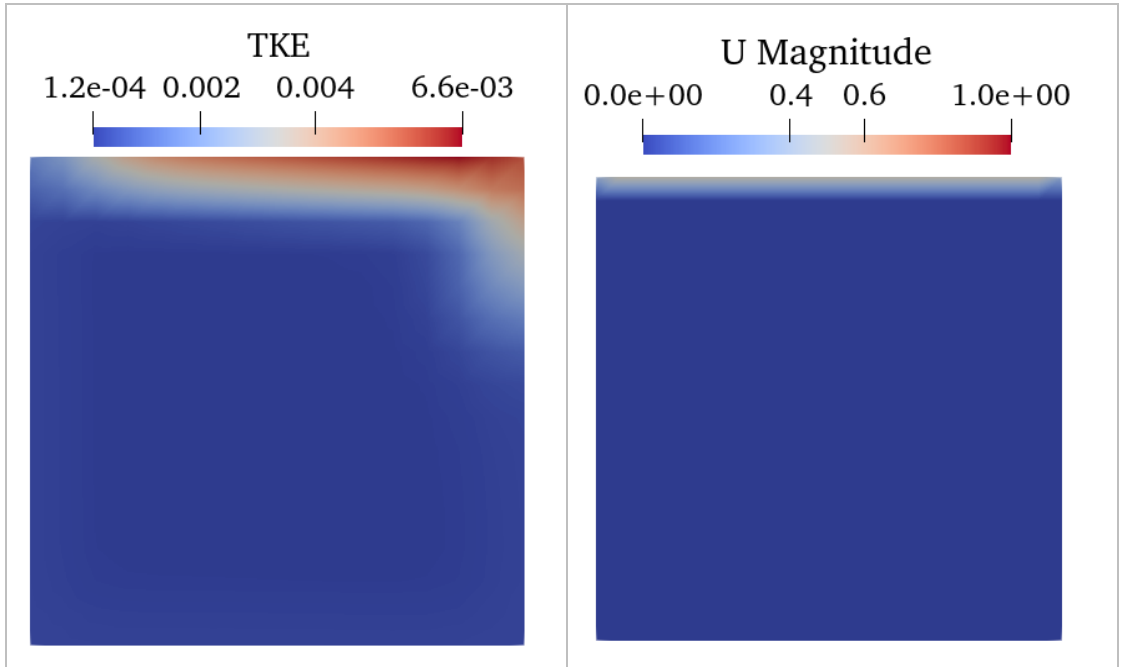

(c)

(d)

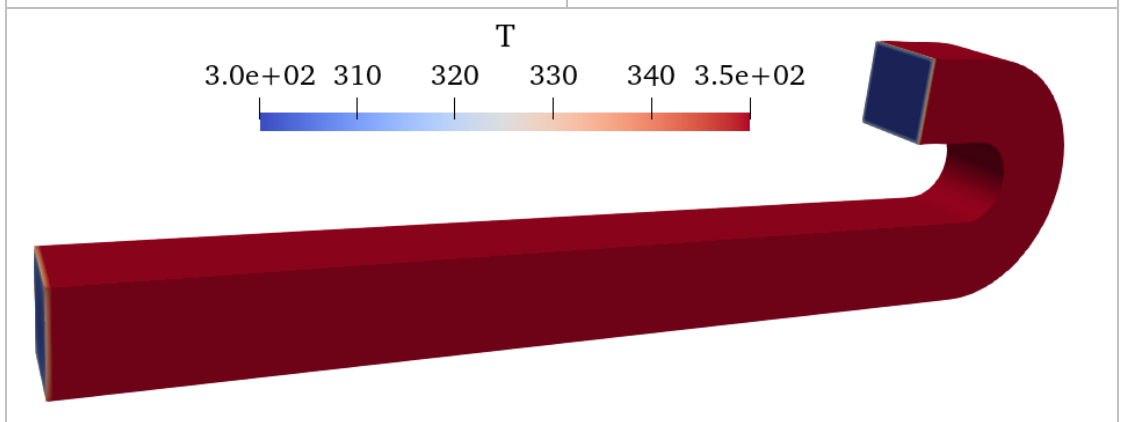

(e)

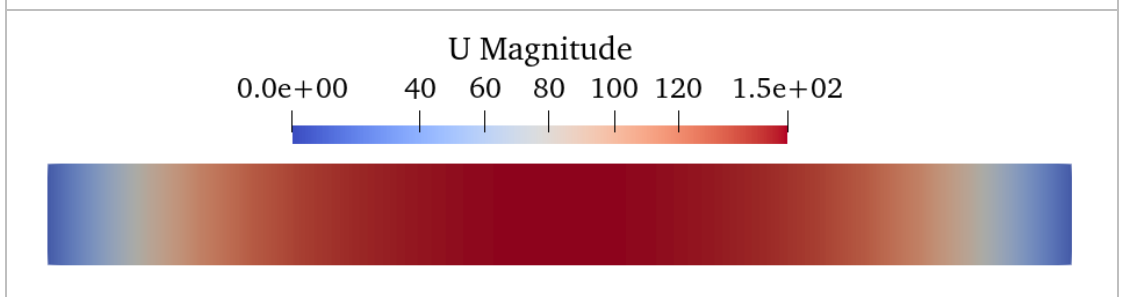

(f)

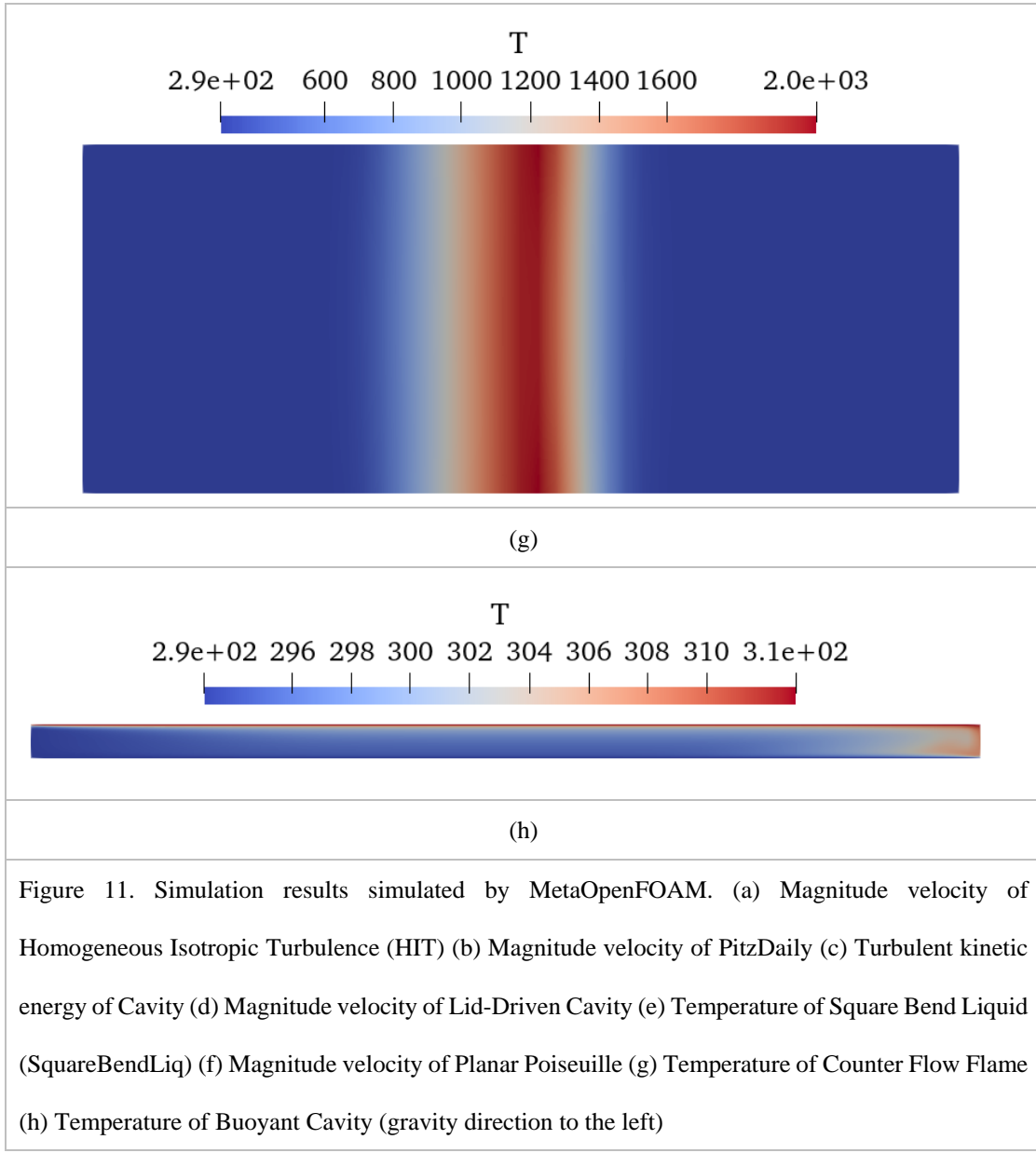

Figure 11. Simulation results simulated by MetaOpenFOAM. (a) Magnitude velocity of Homogeneous Isotropic Turbulence (HIT) (b) Magnitude velocity of PitzDaily (c) Turbulent kinetic energy of Cavity (d) Magnitude velocity of Lid-Driven Cavity (e) Temperature of Square Bend Liquid (SquareBendLiq) (f) Magnitude velocity of Planar Poiseuille (g) Temperature of Counter Flow Flame (h) Temperature of Buoyant Cavity (gravity direction to the left)

**Appendix E. Ablation study**

Table 6. Performance of MetaOpenFOAM when remove Reviewer

|  | Executability | Token Usage | Iteration | Productivity | Pass@1(%) |
|---|---|---|---|---|---|
| **HIT** | 0.6 | 11872.4 | 0 | 32.3 | 0 |
| **PitzDaily** | 1 | 12631.8 | 0 | 22.5 | 0 |
| **Cavity** | 4 | 12848 | 0 | 28.5 | 100 |
| **LidDrivenCavity** | 1.6 | 12681.6 | 0 | 40.8 | 20 |
| **SquareBendLiq** | 4 | 16371.8 | 0 | 27.5 | 100 |
| **PlanarPoiseuille** | 1 | 12578.8 | 0 | 29.9 | 0 |
| **CounterFlowFlame** | 1 | 20781.8 | 0 | 26.1 | 0 |
| **BuoyantCavity** | 0.2 | 25521.3 | 0 | 31.1 | 0 |
| **Average** | 1.7 | 15660.94 | 0 | 29.9 | 27.5 |

Table 7. Performance of MetaOpenFOAM when remove Action Review architecture

| | Executability | Token Usage | Iteration | Productivity | Pass@1(%) |
|---|---|---|---|---|---|
| **HIT** | 0.8 | 165196.0 | 20 | 488.2 | 0 |
| **PitzDaily** | 4 | 18083.2 | 2.1 | 32.1 | 100 |
| **Cavity** | 4 | 12863.6 | 0 | 28.3 | 100 |
| **LidDrivenCavity** | 2.2 | 73685.2 | 12.4 | 241.0 | 40 |
| **SquareBendLiq** | 3.7 | 21529.2 | 2 | 36.4 | 90 |
| **PlanarPoiseuille** | 2.8 | 32691.0 | 5 | 74.0 | 90 |
| **CounterFlowFlame** | 4 | 78864.4 | 4 | 85.4 | 100 |
| **BuoyantCavity** | 2.4 | 151771.6 | 15.8 | 158.4 | 40 |
| **Average** | 3.0 | 69335.5 | 7.7 | 143.0 | 70 |

Table 8. Performance of MetaOpenFOAM when remove RAG

| | Executability | Token Usage | Iteration | Productivity | Pass@1(%) |
|---|---|---|---|---|---|
| **HIT** | 0 | 7000.4 | 20 | 15.8 | 0 |
| **PitzDaily** | 1.4 | 95145.6 | 17.4 | 187.9 | 0 |
| **Cavity** | 0 | 118260 | 20 | 287.0 | 0 |
| **LidDrivenCavity** | 1 | 49751 | 20 | 164.2 | 0 |
| **SquareBendLiq** | 0.5 | 58698 | 20 | 229.7 | 0 |
| **PlanarPoiseuille** | 1.6 | 83655.8 | 16 | 232.8 | 0 |
| **CounterFlowFlame** | 1 | 132260 | 20 | 136.1 | 0 |
| **BuoyantCavity** | 1 | 105996 | 20 | 12.2 | 0 |
| **Average** | 0.8125 | 81345.9 | 19.2 | 158.2 | 0 |

# Appendix F. Detailed discussion of temperature

Table 9. Performance of MetaOpenFOAM when temperature=0.5 in LLM.

|  | Executability | Token Usage | Iteration | Productivity | Pass@1(%) |
|---|---|---|---|---|---|
| **HIT** | 1.8 | 128321.8 | 14.6 | 300.4 | 30 |
| **PitzDaily** | 3.7 | 36209.7 | 4.7 | 58.6 | 90 |
| **Cavity** | 3.7 | 30568.6 | 4.7 | 19.4 | 90 |
| **LidDrivenCavity** | 3.7 | 41296.1 | 6.4 | 44.0 | 90 |
| **SquareBendLiq** | 3.6 | 16449 | 2.9 | 27.6 | 90 |
| **PlanarPoiseuille** | 4 | 40384.9 | 6.2 | 92.0 | 100 |
| **CounterFlowFlame** | 3.1 | 66518.5 | 10.6 | 46.3 | 70 |
| **BuoyantCavity** | 3.7 | 107599.4 | 9.3 | 111.6 | 90 |
| **Average** | 3.4 | 58418.5 | 7.4 | 87.5 | 81.3 |

Table 10. Performance of MetaOpenFOAM when temperature=0.99 in LLM

|  | Executability | Token Usage | Iteration | Productivity | Pass@1(%) |
|---|---|---|---|---|---|
| **HIT** | 1.3 | 161316 | 18 | 407.9 | 10 |
| **PitzDaily** | 3 | 61516.3 | 9.3 | 100.0 | 60 |
| **Cavity** | 3.2 | 38278.4 | 8.6 | 70.3 | 80 |
| **LidDrivenCavity** | 2.2 | 78580.8 | 13 | 199.1 | 40 |
| **SquareBendLiq** | 2 | 48812.4 | 12.8 | 75.8 | 40 |
| **PlanarPoiseuille** | 3.3 | 55319.4 | 8.3 | 126.0 | 80 |
| **CounterFlowFlame** | 2.8 | 165198.2 | 17.6 | 178.0 | 60 |
| **BuoyantCavity** | 1 | 264538 | 20 | 277.6 | 0 |
| **Average** | 2.3 | 109194.9 | 13.5 | 179.3 | 46.3 |

# Appendix G. Modified dataset study

Table 11. Performance of MetaOpenFOAM in modified dataset

|  | Executability | Token Usage | Iteration | Productivity | Pass@1(%) |
|---|---|---|---|---|---|
| **HIT** | 4 | 31386.8 | 4.6 | 91.8 | 100 |
| **PitzDaily** | 4 | 18507.2 | 2.2 | 33.2 | 100 |
| **Cavity** | 4 | 14697.8 | 1.4 | 33.3 | 100 |
| **LidDrivenCavity** | 4 | 31256.6 | 4.4 | 91.9 | 100 |
| **SquareBendLiq** | 2 | 16442.4 | 20 | 27.2 | 0 |
| **PlanarPoiseuille** | 4 | 59351 | 7 | 138.2 | 100 |
| **CounterFlowFlame** | 4 | 35274.8 | 2.8 | 38.0 | 100 |
| **BuoyantCavity** | 3.6 | 180419 | 15 | 180.0 | 80 |
| **Average** | 3.7 | 48417.0 | 7.2 | 79.2 | 85 |